\documentclass[letterpaper, 10 pt, conference]{ieeeconf}

\usepackage{amsmath}
\usepackage{amssymb}
\usepackage{amsfonts}
\usepackage{microtype}
\usepackage{graphicx}
\usepackage{booktabs}
\usepackage{multirow}
\usepackage{stfloats}
\usepackage{float}
\usepackage[utf8]{inputenc}
\usepackage[table]{xcolor}
\usepackage{hyperref}

\hypersetup{
    colorlinks=true,
    linkcolor=blue,
    filecolor=magenta,
    urlcolor=blue,
}

\definecolor{tableblue}{RGB}{230, 242, 250}
\IEEEoverridecommandlockouts
\title{\LARGE \bf
ECHO: Edge-Cloud Humanoid Orchestration for Language-to-Motion Control
}

\author{Haozhe Jia$^{1,2,*}$, Jianfei Song$^{2,*}$, Yuan Zhang$^{3,*}$, Honglei Jin$^{1,3}$, Youcheng Fan$^{1}$, \\
Wenshuo Chen$^{1}$, Wei Zhang$^{2,\dagger}$, and Yutao Yue$^{1,4,\dagger}$%
\thanks{$^*$Equal contribution.}%
\thanks{$^\dagger$Corresponding authors: Wei Zhang and Yutao Yue.}%
\thanks{$^{1}$The Hong Kong University of Science and Technology (Guangzhou).}%
\thanks{$^{2}$LimX Dynamics Technology Co., Ltd.}%
\thanks{$^{3}$Shandong University.}%
\thanks{$^{4}$Institute of Deep Perception Technology, Jiangsu Industrial Technology Research Institute (JITRI).}%
}

\begin{document}

\maketitle
\thispagestyle{empty}
\pagestyle{empty}

\begin{abstract}

We present ECHO, an edge--cloud framework for language-driven whole-body control of humanoid robots. A cloud-hosted diffusion-based text-to-motion generator synthesizes motion references from natural language instructions, while an edge-deployed reinforcement-learning tracker executes them in closed loop on the robot. The two modules are bridged by a compact, robot-native 38-dimensional motion representation that encodes joint angles, root planar velocity, root height, and a continuous 6D root orientation per frame, eliminating inference-time retargeting from human body models and remaining directly compatible with low-level PD control. The generator adopts a 1D convolutional UNet with cross-attention conditioned on CLIP-encoded text features; at inference, DDIM sampling with 10 denoising steps and classifier-free guidance produces motion sequences in approximately one second on a cloud GPU. The tracker follows a Teacher--Student paradigm: a privileged teacher policy is distilled into a lightweight student equipped with an evidential adaptation module for sim-to-real transfer, further strengthened by morphological symmetry constraints and domain randomization. An autonomous fall recovery mechanism detects falls via onboard IMU readings and retrieves recovery trajectories from a pre-built motion library. We evaluate ECHO on a retargeted HumanML3D benchmark, where it achieves strong generation quality (FID 0.029, R-Precision Top-1 0.686) under a unified robot-domain evaluator, while maintaining high motion safety and trajectory consistency. Real-world experiments on a Unitree G1 humanoid demonstrate stable execution of diverse text commands with zero hardware fine-tuning. 

\vspace{0.5em}
\noindent For an online demo and additional video results, please visit our \textbf{\href{https://echo-phi-eight.vercel.app}{Project Page}}.
\end{abstract}

\section{INTRODUCTION}
Natural language is an appealing interface for humanoid robots: it is expressive, low-effort for humans, and naturally describes goals, styles, and constraints. The past few years have witnessed rapid progress in \emph{language-conditioned whole-body control}, driven by stronger vision-language representations, large-scale motion datasets, and improved simulation-to-real (sim2real) reinforcement learning. Recent systems can follow diverse instructions and produce agile full-body behaviors in simulation and, in some cases, on real hardware \cite{shao2025langwbc,wang2025sentinel}. In parallel, modular pipelines that leverage powerful human motion generators (e.g., SMPL/SMPL-X based representations) have enabled language-to-motion generation followed by retargeting and tracking on humanoid platforms \cite{li2026w1,loper2015smpl}. More recently, retargeting-free directions have emerged, for instance by conditioning policies on compact motion latents rather than explicit decoded human poses \cite{li2025roboghost}.

Despite these advances, \emph{deployment-oriented} language-to-humanoid control still faces a fundamental tension between semantic expressivity, real-time closed-loop control, and engineering practicality. 
First, fully end-to-end language-action models (e.g., LangWBC/SENTINEL-style approaches) couple high-level language understanding with low-level control in a single on-board policy \cite{shao2025langwbc,wang2025sentinel}. While elegant, this design often implies that the robot must execute substantial language-action inference under high-frequency control constraints. On resource-limited humanoid computers, maintaining low latency and high control rates can directly compete with running large models, which in turn makes it difficult to scale semantic complexity without sacrificing real-time stability. Moreover, forcing one network to simultaneously solve open-vocabulary instruction grounding \emph{and} contact-rich dynamics increases training and sim2real burdens, and complicates safety enforcement.

Second, modular \emph{human-motion-first} pipelines such as FRoM-W1 generate human whole-body motion in a body model space and then retarget to a specific humanoid \cite{li2026w1,loper2015smpl}. Although they benefit from abundant human motion data, they introduce non-trivial retargeting engineering: morphology mismatch (human body model vs.\ robot kinematics), constraint inconsistencies (joint limits, contacts, balance), and error accumulation across stages. In addition, human body parameterizations often include degrees of freedom that are unnecessary for robot execution, making the generation-to-execution interface less efficient for real-time streaming and tracking.

Third, while retargeting-free latent-guidance approaches \cite{li2025roboghost} mitigate decoding overhead, their latent interfaces are typically co-trained and tightly coupled with downstream policies. This entanglement severely degrades modularity: migrating across robot platforms or integrating with established tracking stacks necessitates complete interface realignment, thereby restricting portability and interpretability.

To address these limitations, we advocate for a strict decoupling of generation and execution in language-to-humanoid pipelines: a generative model should synthesize semantically consistent motion references, whereas a dedicated tracker enforces physical feasibility via closed-loop control. Consequently, we propose \textbf{ECHO}, an inference-time retargeting-free, \emph{edge--cloud} framework for language-directed whole-body control. Specifically, a cloud-deployed diffusion model generates \emph{robot-native} motion references, which are streamed as short-horizon chunks to an on-board, lightweight tracking controller. This distributed architecture circumvents stringent hardware constraints, ensuring low on-board compute overhead while sustaining high-frequency, robust physical execution.

On the \textbf{cloud side}, our generator adopts a 1D convolutional UNet backbone conditioned on language via a frozen CLIP encoder~\cite{radford2021learning} and cross-attention injection. The model is trained with DDPM~\cite{ho2020denoising} using a masked $L_2$ objective and classifier-free guidance~\cite{ho2022classifier}, and uses DDIM~\cite{song2021denoising} at inference to produce high-quality motion in as few as 10 denoising steps, achieving approximately one-second cloud-to-robot latency, balancing generation fidelity and real-time deployment constraints. The training corpus retargets HumanML3D motions~\cite{guo2022generating} (a captioned subset of AMASS~\cite{mahmood2019amass}) to the robot skeleton via General Motion Retargeting (GMR), preserving the original text--motion pairing.
On the \textbf{edge side}, our tracker employs an Asymmetric Actor-Critic Teacher-Student pipeline~\cite{pinto2017asymmetric}: a teacher policy is first trained with access to privileged physical states under PPO~\cite{schulman2017proximal}. A student adaptation module is then trained with Evidential Deep Regression~\cite{amini2020deep} to handle sim-to-real uncertainty and distill the teacher's behavior using only proprioceptive history. A morphological symmetry loss is further introduced to eliminate asymmetric gait artifacts during training.

\begin{figure*}[htbp]
    \centering
    \includegraphics[width=\textwidth]{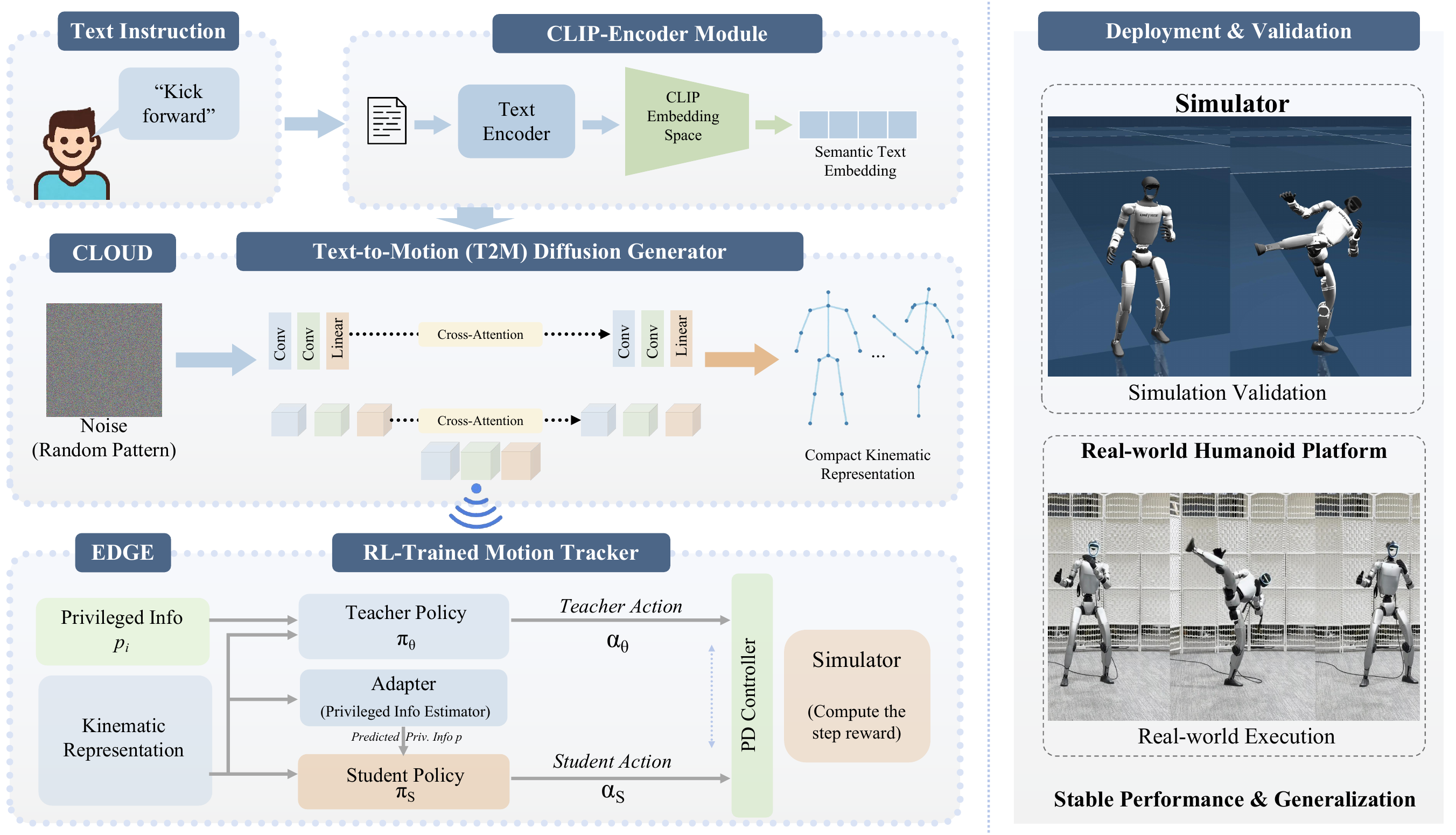} 
    \caption{
        \textbf{Overview of the proposed framework:} The system features a Cloud-Edge decoupled deployment. The Cloud module utilizes a Diffusion Generator to synthesize motion from text instructions via CLIP encoding. The Edge module employs an RL-trained Student Policy that tracks targets using estimated privileged information. The resulting actions are executed via a PD controller for stable humanoid motion in the real world.
    }
    \label{fig:teaser_overview}
\end{figure*}

A key element bridging both modules is a compact 38D velocity-based robot motion representation that directly matches the humanoid execution interface. Each frame is a 38-dimensional vector composed of: (i) $29$D joint positions (robot joint angles), (ii) $2$D root planar velocity in an aligned global frame, (iii) $1$D root height, and (iv) $6$D continuous root rotation representation to avoid discontinuities \cite{zhou2019continuity}. Motions are generated at $50$ FPS and transmitted to the on-board tracker via a persistent WebSocket connection. This representation avoids explicit global root positions (reducing drift and improving robustness), removes unnecessary human-body degrees of freedom, and provides a stable target for a receding-horizon tracker to follow under contacts and disturbances.

We validate the proposed framework in extensive simulation experiments and on a real humanoid robot. Results show that our system produces instruction-consistent, temporally coherent whole-body behaviors while improving deployment efficiency and robustness, compared with representative end-to-end and retargeting-based baselines.

Our main contributions are:
\begin{itemize}
    \setlength{\itemsep}{0pt}
    \setlength{\parskip}{0pt}
    \setlength{\parsep}{0pt}
    \setlength{\topsep}{0pt}
    \item A deployment-oriented \textbf{edge--cloud} language-to-humanoid control paradigm that streams diffusion-generated motion references from the cloud and executes them with an on-board tracker in closed loop.
    \item A \textbf{robot-native 38D velocity-based} motion interface that is retargeting-free at inference time, streamable, and tracker-friendly, bridging generative motion planning and real-time humanoid execution.
    \item Comprehensive evaluations in simulation and \textbf{real-robot} experiments, along with two novel robot-centric metrics: \textbf{Motion Safety Score (MSS)} and \textbf{Root Trajectory Consistency (RTC)}, which quantify hardware constraint compliance and path-shape fidelity, two properties critical for safe deployment that standard text-to-motion benchmarks do not cover.
\end{itemize}

\section{RELATED WORK}
\subsection{Motion Generation}
3D human motion generation has evolved from early GAN-based \cite{ahn2018text2action} and AE-based \cite{ahuja2019language2pose} deterministic mappings to high-fidelity probabilistic synthesis. To improve diversity, VAE-based frameworks like TEMOS \cite{petrovich2022temos} aligned text and motion in joint latent spaces. Recently, diffusion models such as MDM \cite{tevet2023human} and MotionDiffuse \cite{zhang2024motiondiffuse} have set new benchmarks via iterative denoising, while MLD \cite{chen2023executing} enhances efficiency through latent space diffusion. Further advancements continue to improve temporal alignment \cite{Chen_2025}, frequency-domain robustness for robotic applications \cite{chen2025freet2mrobusttexttomotiongeneration}, and fundamental diffusion properties \cite{chen2025polarisprojectionorthogonalsquaresrobust, ning2025dctdiffintriguingpropertiesimage}. Despite these advances, purely kinematic methods often struggle with physical artifacts such as foot sliding and penetration. While models like PhysDiff \cite{yuan2023physdiff} integrate physics-based projection into the diffusion loop to mitigate these issues, they often incur significant computational overhead. Beyond standalone T2M synthesis, recent deployment-oriented systems further couple motion generation with physically executable humanoid control, either through modular pipelines with retargeting/tracking \cite{li2026w1,xie2026textop} or by bypassing explicit decoding/retargeting via latent-guided control \cite{li2025roboghost}.

Different from previous monolithic or purely kinematic approaches, our framework facilitates an integrated pipeline that decouples instruction parsing, T2M diffusion generation, and RL-based physical tracking. The diffusion component ensures semantic alignment and temporal consistency in the synthesized motion, while the tracking module translates these kinematic references into dynamically feasible whole-body maneuvers for robust execution on humanoid platforms.

\subsection{Whole Body Motion Tracking}
The core objective of motion tracking is to faithfully execute given kinematic references. Early approaches relied heavily on trajectory optimization \cite{kuindersma2016optimization}, while Deep Reinforcement Learning (DRL) popularized data-driven imitation, utilizing task-specific rewards to minimize tracking errors for individual skills \cite{peng2018deepmimic}. To overcome the scalability bottlenecks of this ``one-skill-one-policy'' paradigm, unified tracking controllers emerged. By leveraging adversarial objectives \cite{peng2021amp}, a single policy can robustly track extensive motion databases without per-task retraining.

Recently, the scope of tracking has expanded to executing dynamically synthesized trajectories, particularly those from language-conditioned motion generators \cite{shao2025langwbc,wang2025sentinel}. In such hierarchical pipelines, a universal low-level tracking policy operates in a closed loop to follow upstream kinematic references \cite{li2026w1,xie2026textop}. To minimize staging latency during tracking, retargeting-free approaches condition the tracker directly on language-grounded latents rather than explicit poses \cite{li2025roboghost}. Finally, deploying these universal tracking policies onto physical humanoids relies fundamentally on robust sim-to-real transfer to bridge the dynamics gap.
\section{METHOD}

As illustrated in Fig.~\ref{fig:teaser_overview}, the proposed framework decouples language understanding from real-time control by distributing computation across cloud and edge. A diffusion-based text-to-motion generator running on a remote server synthesizes robot-native motion references from natural language instructions. These reference sequences, represented in the compact 38D format described below, are then streamed to the on-board controller. The edge-deployed tracking policy executes the received references in closed loop, maintaining high control rates and robustness under contact and external disturbances. This separation of concerns enables flexible scaling of semantic complexity on the cloud side while preserving low-latency, physically grounded execution on the robot. In the following, we detail each component.

\subsection{Robot-Skeleton Motion Representation.}
To enable stable diffusion-based generation, we transform raw robot kinematics into a compact, learning-friendly feature space. Given a robot with $N_{dof}=29$ actuators, the frame-wise state $\mathbf{m}_t \in \mathbb{R}^{38}$ is defined as:
\begin{equation}
\mathbf{m}_t = \Big[\ \mathbf{q}_t,\ \mathbf{v}^{root}_t,\ h^{root}_t,\ \mathbf{R}^{6D}_t\ \Big],
\label{eq:repr38}
\end{equation}
where $\mathbf{q}_t \in \mathbb{R}^{29}$ are joint angles, $\mathbf{v}^{root}_t \in \mathbb{R}^{2}$ is the planar root velocity, $h^{root}_t \in \mathbb{R}$ is root height, and $\mathbf{R}^{6D}_t \in \mathbb{R}^6$ represents root orientation.

\smallskip
\noindent\textbf{Why Velocity for Root XY?}
A critical design choice is representing the planar root translation as frame-wise velocity $\mathbf{v}^{root}_t = \mathbf{p}^{xy}_t - \mathbf{p}^{xy}_{t-1}$, while keeping other dimensions absolute. This is because absolute XY coordinates are \emph{unbounded} and non-periodic, making them difficult for neural networks to predict directly. Learning absolute positions often leads to mode collapse or ``teleportation'' artifacts where the robot jumps discontinuously. In contrast, joint angles (physically limited), root height (ground-constrained), and rotations (cyclic) naturally lie on bounded manifolds. By predicting local velocity, we ensure the diffusion model learns consistent locomotion dynamics rather than memorizing global locations.

\smallskip
\noindent\textbf{Why 6D Rotation?}
We employ the continuous 6D representation~\cite{zhou2019continuity} for root orientation instead of quaternions. The primary limitation of quaternions is the strict \emph{unit norm constraint} ($\| \mathbf{q}_{rot} \|_2 = 1$). Neural networks struggle to output strictly normalized vectors, and enforcing normalization post-hoc can disrupt gradient flow. The 6D representation is formed by the first two columns of the rotation matrix and requires no such constraint. Any $6D$ vector can be mapped to a valid rotation through Gram–Schmidt orthogonalization, which significantly stabilizes the training process.

\smallskip
\noindent\textbf{Control-Centric Advantages.}
This joint-space representation ($\mathbf{q}_t$) is inherently compatible with robotic control. Unlike Cartesian-based methods that require Inverse Kinematics (IK), our generated sequences directly map to the robot's actuators, ensuring kinematic feasibility and enabling seamless tracking by low-level PD controllers.

\subsection{Cloud: Text-to-Motion Generator}
We adopt a diffusion-based generative model to synthesize full-body motion sequences conditioned on natural language instructions. The model directly operates on the robot-native 38D representation, eliminating the need for inference-time retargeting from human body models.

\smallskip
\noindent\textbf{Model Architecture.}
The generative backbone employs a 1D convolutional UNet architecture with residual temporal blocks and adaptive group normalization (AdaGN). Text instructions are first encoded by a frozen CLIP ViT-B/32 vision-language model \cite{radford2021learning}, whose output is refined by a lightweight Transformer encoder to produce a $d_{text}$-dimensional language latent $\mathbf{z}_{text}$. This latent is injected into each UNet block via cross-attention layers, enabling the network to condition motion generation on semantic instruction features. The UNet structure with down/up-sampling paths provides efficient multi-scale temporal modeling for motion sequences.

We also implement a Transformer-based encoder-decoder variant (referred to as ECHO-Transformer in Table~\ref{tab:main_results}) for architectural comparison. The model is trained to reverse a forward diffusion process that gradually corrupts a clean motion sequence $\mathbf{M} = \{\mathbf{m}_1, \dots, \mathbf{m}_T\}$ with Gaussian noise. Given a text prompt $c$, the network learns to predict the denoised sequence $\hat{\mathbf{M}}_0$ from a noisy observation $\mathbf{M}_t$ at timestep $t$.

\smallskip
\noindent\textbf{Training.}
We follow the DDPM objective \cite{ho2020denoising} with $T$ diffusion timesteps and a linear noise schedule. At each training step, a random timestep $t \sim \text{Uniform}(1, T)$ is sampled, and the network is tasked with directly predicting the clean motion $\mathbf{M}_0$ from the noised input $\mathbf{M}_t$. The loss function is a masked $L_2$ reconstruction error, computed only over valid frames to handle variable-length sequences:
\begin{equation}
    \mathcal{L}_{diffusion} = \mathbb{E}_{t, \mathbf{M}_0} \left[ \| \mathbf{M}_0 - \hat{\mathbf{M}}_0(\mathbf{M}_t, t, c) \|_2^2 \odot \mathbf{mask} \right],
\end{equation}
where $\mathbf{mask}$ indicates valid frame positions and $\mathbf{M}_t = \sqrt{\bar\alpha_t}\mathbf{M}_0 + \sqrt{1-\bar\alpha_t}\,\epsilon$ with $\epsilon \sim \mathcal{N}(0, \mathbf{I})$.

To enable classifier-free guidance \cite{ho2022classifier}, we randomly drop the text condition during training with probability $p_{uncond}$, replacing $\mathbf{z}_{text}$ with a learned null embedding. At inference time, the model steers generation beyond the conditional prediction to enhance instruction adherence.

The training corpus retargets HumanML3D motions~\cite{guo2022generating} (a captioned subset of AMASS~\cite{mahmood2019amass}) to the target robot skeleton using General Motion Retargeting (GMR), which preserves motion naturalness while adapting to the robot's kinematic constraints; the original text--motion pairing is retained. All motion features are z-score normalized per dimension to stabilize training. We optimize the model using AdamW \cite{loshchilov2018decoupled} with learning rate $\eta$, and maintain an Exponential Moving Average (EMA) of the weights with decay $\beta_{ema}$ for stable generation at inference.

\begin{figure*}[t]
\centering


\newcommand{\rowh}{2.2cm}
\newcommand{\colgap}{12pt}
\newcommand{\rowgap}{6pt}
\newlength{\colw}
\setlength{\colw}{\dimexpr(\linewidth-\colgap)/2\relax}
\newcommand{\cellimg}[1]{%
  \includegraphics[width=\colw, height=\rowh]{#1}%
}

\begin{tabular*}{\linewidth}{@{}p{\colw}@{\hspace{\colgap}}p{\colw}@{}}
\multicolumn{1}{c}{\textbf{Sim (MuJoCo)}} & \multicolumn{1}{c}{\textbf{Real}} \\[4pt]

\cellimg{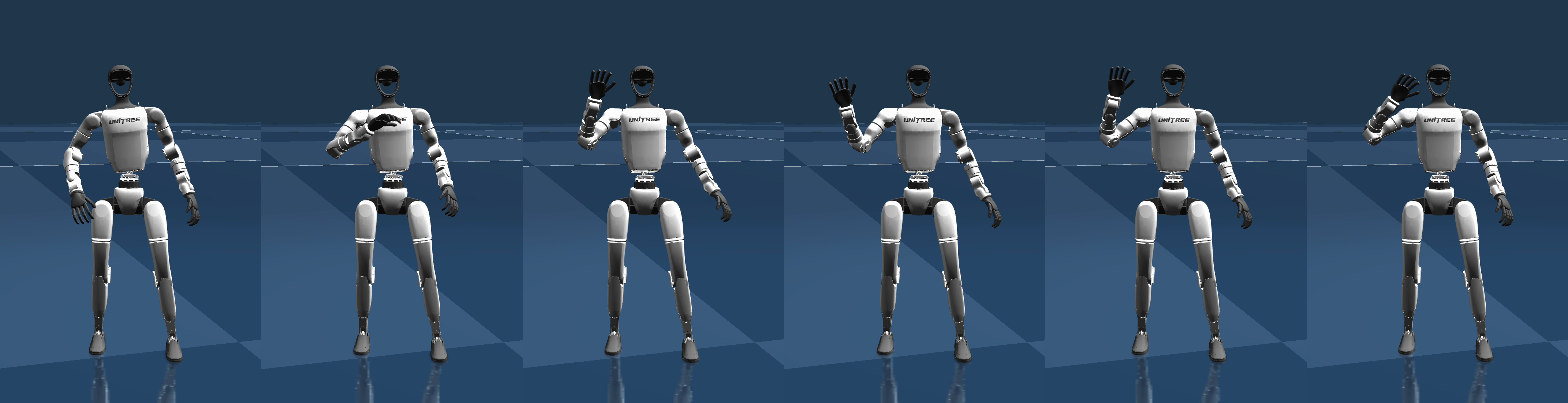} &
\cellimg{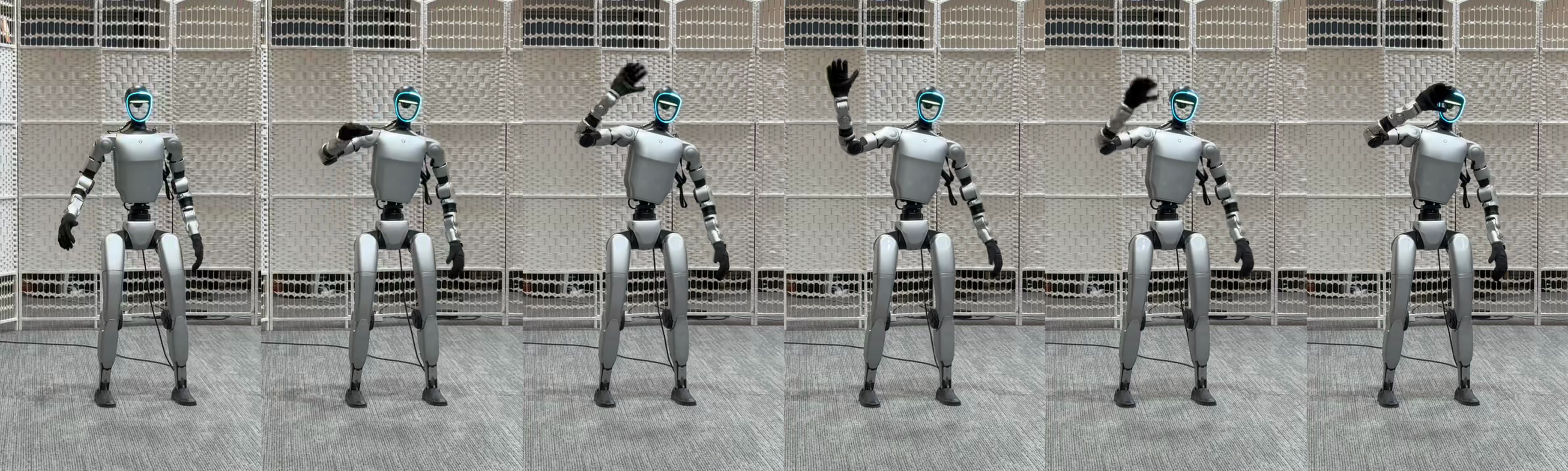} \\
\multicolumn{2}{c}{\footnotesize\textit{``A man is waving his right hand.''}} \\[\rowgap]
\cellimg{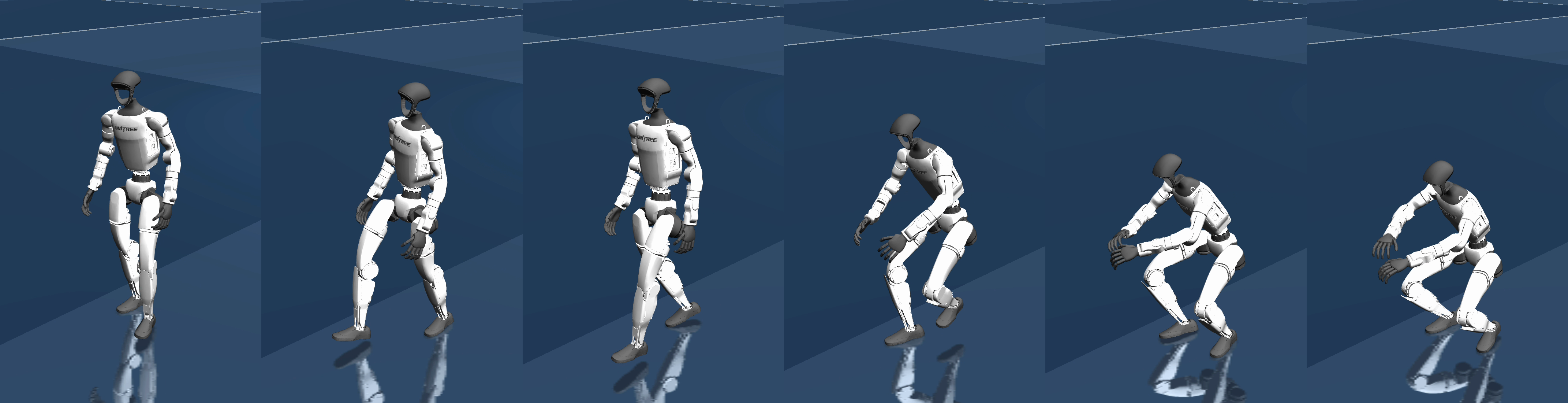} &
\cellimg{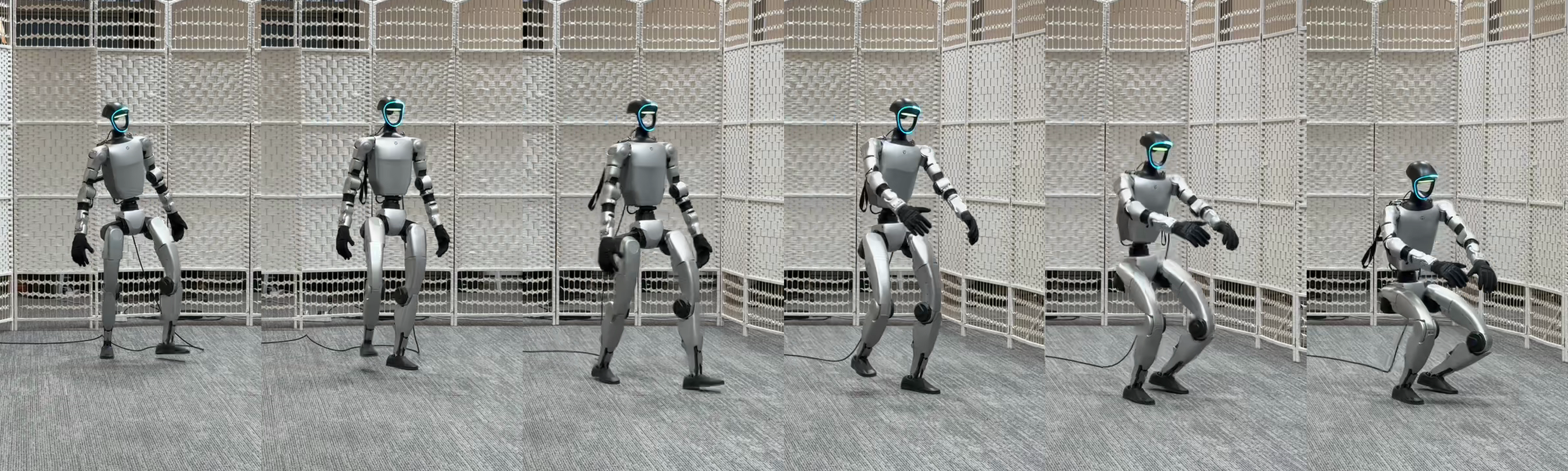} \\
\multicolumn{2}{c}{\footnotesize\textit{``A man walks forward then squats.''}} \\[\rowgap]
\cellimg{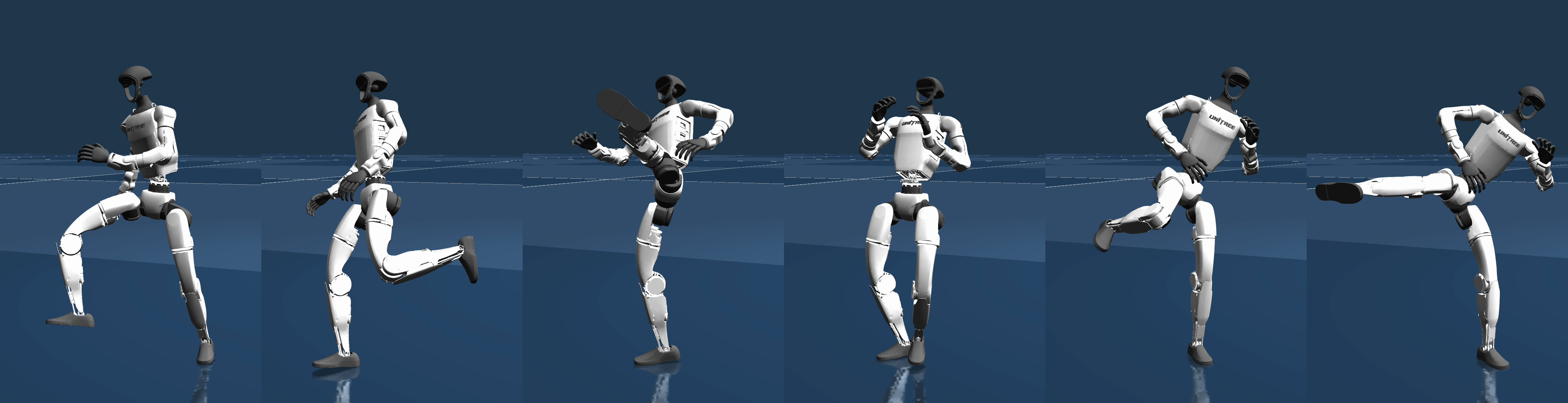} &
\cellimg{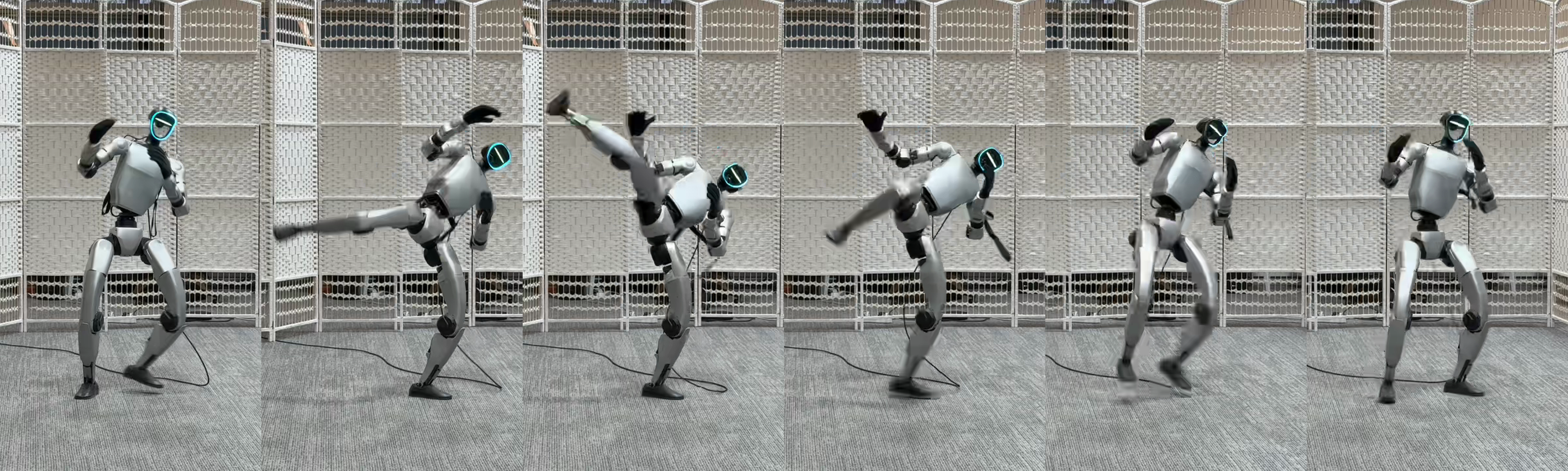} \\
\multicolumn{2}{c}{\footnotesize\textit{``A man kicks forward.''}} \\
\end{tabular*}

\caption{\textbf{Sim-to-Real Results:} Validation of robust tracking performance from simple gestures to dynamic maneuvers.}
\label{fig:sim2real}
\end{figure*}

\smallskip
\noindent\textbf{Inference.}
At deployment, we employ DDIM \cite{song2021denoising} as our sampling strategy, which was selected through extensive ablation studies comparing DDPM, DDIM, and DPM-Solver schedulers (see Table~\ref{tab:ablation_final_no_rtc} for details). DDIM enables efficient generation with only 10 denoising steps while maintaining high fidelity, reducing latency compared to the full 1000-step DDPM sampling. Classifier-free guidance is applied with scale $s=2.5$ to enhance semantic fidelity:
\begin{equation}
    \hat{\mathbf{M}}_0^{guided} = \hat{\mathbf{M}}_0^{uncond} + s \cdot (\hat{\mathbf{M}}_0^{cond} - \hat{\mathbf{M}}_0^{uncond}).
\end{equation}
The generator is deployed as a persistent inference server on cloud infrastructure, serving text-driven motion requests in real time (communication details in Section~\ref{sec:deployment}).

\subsection{Tracker}
\label{sec:tracker}
To execute precise whole-body motions, we build upon the teacher-student tracking architecture from prior work \cite{lu2025gentlehumanoid}. 

The teacher phase pre-trains an asymmetric actor-critic model \cite{pinto2017asymmetric}. First, an encoder $E_\phi$ compresses high-dimensional privileged states into latent features $z_t = E_\phi(x_t)$. The teacher policy $\pi_\theta(a_t | o_t, z_t)$ then maps these features alongside proprioceptive observations $o_t$ to joint targets. To reduce variance, the critic $V_\psi(o_t, x_t)$ leverages the complete environmental state. We optimize the policy using Proximal Policy Optimization (PPO) \cite{schulman2017proximal} with Generalized Advantage Estimation \cite{schulman2015high}.

The composite reward comprises a tracking term $R_{track}$ and a regularization term $R_{reg}$. Following DeepMimic~\cite{peng2018deepmimic}, $R_{track}$ aggregates exponential squared errors across joint positions and velocities:
\begin{equation}
    R_{track} = \sum_{k \in \{\text{pos}, \text{vel}\}}w_k \exp \left( - \frac{\| \mathbf{x}_k - \mathbf{x}^*_k \|^2}{\sigma_k} \right)
\end{equation}
where $\mathbf{x}_k$ and $\mathbf{x}^*_k$ denote the simulated actual state and the desired reference state for feature $k$, respectively. $w_k$ represents the weight coefficient for each specific feature, and $\sigma_k$ is the scale factor controlling the sensitivity of the reward kernel.

Furthermore, pure kinematic tracking often leads to destructive interaction forces. Therefore, strict physical feasibility penalties are introduced, including joint acceleration limits, motor torque soft limits, and a quadratic penalty for excessive landing impact forces 
$r_{impact} = - \sum (v^{\downarrow}_{z})^2 \cdot\mathbb{I}_{contact}$. 
Concurrently, by incorporating a dense feet air-time reward, the robot is guided to form smooth swing-stance phase transitions \cite{rudin2022learning}.

To deploy the policy to the real world without privileged information $x_t$, the student adaptation phase trains an adaptation module $A_\omega$ using proprioceptive history $H_t = \{o_{t-H}, \dots, o_t\}$. To mitigate sim-to-real uncertainties, we model hidden state inference via evidential deep regression \cite{amini2020deep}. Specifically, $A_\omega(H_t)$ outputs a Normal-Inverse-Gamma (NIG) distribution parameterized by $\boldsymbol{\phi}_t = (\mu_z, \nu, \alpha, \beta)$, capturing both the estimated latent $z_t$ and its predictive variance. We optimize this module using a negative log-likelihood (NLL) loss augmented with an evidence regularizer:
\begin{equation}
    \mathcal{L}_{adapt}(\omega) = \mathcal{L}_{NLL}(z_t | \mu_z, \nu, \alpha, \beta) + \lambda_{\text{reg}} \| z_t - \mu_z \| (2\nu + \alpha)
\end{equation}
This regularizer strictly penalizes overconfident erroneous predictions. The student policy $\pi_S(a_t | o_t, \mu_z)$ then recovers the teacher's behavior via behavioral cloning, followed by low-rate PPO fine-tuning.

To accelerate learning and eliminate asymmetric gaits (e.g., limping) \cite{su2024leveraging,siekmann2021sim}, we embed a morphological symmetry prior directly into the PPO update. Let $T_{obs}$ and $T_{act}$ denote mirror operators for the observation and action spaces. We penalize asymmetric policy outputs via:
\begin{equation}
\begin{split}
    \mathcal{L}_{sym} =\ & c_{\mu} \| \mu_\theta(o_t) - T_{act}(\mu_\theta(T_{obs}(o_t))) \|^2 \\
    & + c_{\sigma} \| \sigma_\theta(o_t) - T_{act}(\sigma_\theta(T_{obs}(o_t))) \|^2
\end{split}
\end{equation}
where $c_{\mu}$ and $c_{\sigma}$ are weighting coefficients. Finally, we apply domain randomization to physical parameters (e.g., link masses, friction, actuator dynamics) during simulation rollouts to further bridge the reality gap.

\subsection{Deployment}
\label{sec:deployment}
The trained generator and tracker are integrated into a distributed edge-cloud deployment architecture designed for practical humanoid control.

\smallskip
\noindent\textbf{Edge-Cloud Communication.}
The cloud-side inference server receives natural language instructions via a WebSocket connection, generates the corresponding motion sequence using the text-to-motion diffusion model, and transmits the result back to the robot. The server maintains a persistent connection, allowing successive instructions to be issued without re-initialization. The compact 38D representation minimizes network bandwidth, making the system viable even under moderate wireless latency.

\smallskip
\noindent\textbf{On-Board Execution.}
The student tracking policy, trained via the Teacher-Student distillation pipeline described in Section~\ref{sec:tracker}, is exported to a lightweight inference format optimized for embedded deployment. At each control cycle, the policy receives the current proprioceptive state and the received motion reference, computes target joint positions, and remaps them to the robot's actuation interface. To mitigate discretization artifacts and sudden reference transitions, an exponential moving average filter is applied to smooth action outputs before dispatching them to the low-level PD controller. This filtering preserves responsiveness while preventing torque spikes that could destabilize contact.

\smallskip
\noindent\textbf{Sim-to-Real Transfer.}
Policies are first validated in a MuJoCo-based simulation environment (sim2sim) before hardware deployment. We then transfer the student policy directly to a physical Unitree G1 (29-DoF EDU) humanoid robot without any hardware fine-tuning; the extensive domain randomization described in Section~\ref{sec:tracker} proves critical in bridging the reality gap.

\smallskip
\noindent\textbf{Fall Recovery.}
To enhance real-world robustness, we implement an IMU-triggered autonomous fall recovery mechanism. It utilizes a two-stage retrieval procedure that first filters candidates by gravity alignment and then ranks them by joint configuration similarity to extract and execute the optimal recovery trajectory from a pre-built library.

\section{EXPERIMENTS AND RESULTS}

In this section, we evaluate the proposed ECHO framework from two complementary perspectives: 

(1) Generative Quality, assessing the fidelity, diversity, and semantic alignment of the generated motions from text prompts; 

(2) Tracking Success Rate in Real-World Deployment, verifying the physical feasibility and robustness of our tracker on hardware.

\subsection{Evaluation of Generation Quality}

\noindent\textbf{Metrics.}
We evaluate kinematic reference quality using standard HumanML3D benchmarks via a robot-domain \textbf{MoCLIP} evaluator. We further propose two robot-centric metrics: \textbf{Motion Safety Score (MSS)} for hardware constraint compliance, and \textbf{Root Trajectory Consistency (RTC)} for path-shape fidelity. Full definitions are in Appendix~\ref{sec:metrics}.
\begin{table*}[t]
\centering
\caption{Quantitative comparison of text-to-motion generation methods on the Retargeted HumanML3D. All baselines are re-evaluated under the same MoCLIP on the retargeted test split. $\pm$ denotes 95\% confidence intervals. \textbf{Bold}: best overall; \underline{underline}: best among prior works.}
\label{tab:main_results}
\resizebox{\textwidth}{!}{
\begin{tabular}{l ccc ccc cc}
\toprule
\multirow{2}{*}{Method} & \multicolumn{3}{c}{R-Precision $\uparrow$} & \multirow{2}{*}{FID$\downarrow$} & \multirow{2}{*}{Div.$\uparrow$} & \multirow{2}{*}{MM Dist.$\downarrow$} & \multirow{2}{*}{MSS $\uparrow$} & \multirow{2}{*}{RTC $\uparrow$} \\
\cmidrule(lr){2-4}
& Top 1 & Top 2 & Top 3 & & & & & \\
\midrule
Ground Truth & $0.774^{\pm.004}$ & $0.902^{\pm.002}$ & $0.943^{\pm.001}$ & $0.002$ & $1.375^{\pm.003}$ & - & $0.546^{\pm.000}$ & - \\
\midrule
MDM \cite{tevet2023human} & $0.578^{\pm.004}$ & $0.755^{\pm.004}$ & $0.832^{\pm.004}$ & $0.073^{\pm.001}$ & $1.362^{\pm.004}$ & $0.651^{\pm.017}$ & $0.527^{\pm.001}$ & $\underline{0.442^{\pm.002}}$ \\
TM2T \cite{guo2022tm2t}& $0.461^{\pm.003}$ & $0.612^{\pm.004}$ & $0.695^{\pm.004}$ & $0.183^{\pm.001}$ & $1.300^{\pm.004}$ & $0.561^{\pm.017}$ & $\mathbf{\underline{0.542^{\pm.001}}}$ & $0.413^{\pm.002}$ \\
MotionDiffuse \cite{zhang2024motiondiffuse} & $0.269^{\pm.003}$ & $0.477^{\pm.003}$ & $0.696^{\pm.002}$ & $0.213^{\pm.004}$ & $1.310^{\pm.006}$ & $1.226^{\pm.082}$ & $0.323^{\pm.002}$ & $0.195^{\pm.008}$ \\
StableMofusion \cite{huang2024stablemofusion}& $\underline{0.683^{\pm.003}}$ & $\underline{0.843^{\pm.003}}$ & $\underline{0.905^{\pm.002}}$ & $\underline{0.037^{\pm.000}}$ & $\underline{1.368^{\pm.004}}$ & $\underline{0.414^{\pm.013}}$ & $0.405^{\pm.001}$ & $0.421^{\pm.002}$ \\
\midrule
ECHO-Transformer & $0.668^{\pm.004}$ & $0.829^{\pm.003}$ & $0.895^{\pm.002}$ & $0.061^{\pm.001}$ & $\mathbf{1.370^{\pm.005}}$ & $0.375^{\pm.013}$ & $0.511^{\pm.001}$ & $\mathbf{0.505^{\pm.002}}$ \\
\rowcolor{tableblue} \textbf{ECHO-UNet (Ours)} & $\mathbf{0.686^{\pm.004}}$ & $\mathbf{0.845^{\pm.002}}$ & $\mathbf{0.906^{\pm.003}}$ & $\mathbf{0.029^{\pm.001}}$ & $1.366^{\pm.004}$ & $\mathbf{0.343^{\pm.011}}$ & $0.484^{\pm.001}$ & $0.493^{\pm.002}$ \\
\bottomrule
\end{tabular}
}
\end{table*}

\begin{table}[htbp]
\centering
\caption{Ablation study on inference configurations (scheduler, denoising steps, and CFG scale). All variants use the ECHO-UNet backbone. \textbf{Bold}: best result within each group.}
\label{tab:ablation_final_no_rtc}
\resizebox{\columnwidth}{!}{
\begin{tabular}{l ccccc}
\toprule
Variant & R1 $\uparrow$ & FID $\downarrow$ & MSS $\uparrow$ & RTC $\uparrow$ & Time (s) $\downarrow$ \\
\midrule
\multicolumn{6}{l}{\cellcolor{gray!15}\textbf{Scheduler}} \\
\quad DPM-Solver & 0.643 & 0.040 & 0.466 & 0.431 & 1.02 \\
\quad DDPM       & 0.686 & 0.030 & \textbf{0.492} & \textbf{0.498} & 1.02 \\
\quad DDIM       & \textbf{0.686} & \textbf{0.029} & 0.484 & 0.493 & 1.02 \\
\multicolumn{6}{l}{\cellcolor{gray!15}\textbf{Denoising Steps (DDIM)}} \\
\quad 5 steps   & \textbf{0.697} & 0.029 & 0.481 & \textbf{0.508} & \textbf{0.50} \\
\quad 10 steps  & 0.686 & \textbf{0.029} & \textbf{0.484} & 0.493 & 1.02 \\
\quad 20 steps  & 0.691 & 0.029 & 0.483 & 0.480 & 2.05 \\
\quad 50 steps  & 0.690 & 0.031 & 0.480 & 0.466 & 5.16 \\
\quad 100 steps & 0.685 & 0.031 & 0.479 & 0.458 & 10.33 \\
\multicolumn{6}{l}{\cellcolor{gray!15}\textbf{CFG Scale (DDIM, 10 steps)}} \\
\quad CFG $=$ 1.0 & 0.630 & 0.039 & \textbf{0.530} & \textbf{0.582} & 1.02 \\
\quad CFG $=$ 2.5 & \textbf{0.686} & \textbf{0.029} & 0.484 & 0.493 & 1.02 \\
\quad CFG $=$ 5.0 & 0.663 & 0.047 & 0.290 & 0.378 & 1.02 \\
\midrule
\rowcolor{blue!10} \textbf{ECHO (Ours)} & \textbf{0.686} & \textbf{0.029} & \textbf{0.484} & \textbf{0.493} & \textbf{1.02} \\
\bottomrule
\end{tabular}
}
\end{table}
We first analyze the quality of the generated reference motions prior to the tracking stage. As shown in Table~\ref{tab:main_results}, ECHO-UNet achieves the best overall performance, outperforming prior baselines in FID (0.029, a $21.6\%$ improvement over StableMofusion), MM Dist. (0.343), and R-Precision (Top-1: 0.686). Notably, it attains the highest RTC (0.493) among all prior baselines, confirming superior trajectory consistency for downstream tracking. While ECHO-Transformer yields a marginally higher RTC, the significant performance gap in semantic and diversity metrics validates the UNet backbone's advantage in capturing multi-scale temporal structures.

\smallskip
\noindent\textbf{Ablation Study.}
Table~\ref{tab:ablation_final_no_rtc} presents ablations on three inference hyperparameters. \textit{(i) Scheduler}: DDIM yields the best semantic generation quality, achieving the lowest FID while maintaining competitive R-Precision, MSS, and RTC. \textit{(ii) Denoising Steps}: Performance largely saturates within 5--10 steps. We adopt 10 steps as the default setting, which preserves strong generation quality while keeping the cloud-to-edge latency at about 1.02\,s. \textit{(iii) CFG Scale}: Lower guidance scales ($s=1.0$) favor physical compliance, as reflected by higher MSS and RTC, whereas $s=2.5$ provides the strongest semantic alignment, achieving the best R-Precision and FID. Performance degrades at $s=5.0$, so we select $s=2.5$ as the best trade-off between instruction fidelity and physical safety.

\subsection{Real-World Deployment}
We conduct real-world experiments across a diverse set of tasks. A trial is counted as a ``Success'' only if the humanoid completes the instructed behavior without losing balance, falling, or violating joint torque limits during execution.

Fig.~\ref{fig:sim2real} presents a qualitative comparison between the simulated kinematics in MuJoCo and the real-world execution. The evaluated tasks encompass varying degrees of whole-body dynamics, ranging from upper-body gestures (``A man is waving his right hand.'') to highly dynamic lower-body maneuvers and postural transitions (``A man walks forward then squats.'', ``A man kicks forward.''). The policy successfully bridges the reality gap; the physical robot closely tracks the simulated reference motions without exhibiting instability, effectively handling the unmodeled real-world dynamics and hardware constraints inherent in agile tasks.

Quantitative metrics for text-driven command execution are detailed in Table~\ref{tab:tracking_eval}. For each command, we conducted 20 independent trials, reporting the Success Rate (Succ), Global Mean Per Joint Position Error ($E_\text{g-mpjpe}$), and Local Mean Per Joint Position Error ($E_\text{mpjpe}$).

\begin{table}[H]
    \centering
   \caption{Real-world tracking performance over 20 trials. $E_{\text{g-mpjpe}}$ and $E_{\text{mpjpe}}$ are in mm; $\pm$ denotes standard deviation.}
    \label{tab:tracking_eval}
    \resizebox{\columnwidth}{!}{
    \begin{tabular}{l ccc}
        \toprule
        Text Command & Succ $\uparrow$ & $E_\text{g-mpjpe}$ (mm) $\downarrow$ & $E_\text{mpjpe}$ (mm) $\downarrow$ \\
        \midrule
        ``punch''                       & $20/20$ & $287.420^{\pm 50.200}$ & $32.274^{\pm 1.764}$ \\
        ``wave right hand''             & $20/20$ & $89.681^{\pm 6.345}$   & $22.780^{\pm 0.619}$ \\
        ``strum guitar with left hand'' & $20/20$ & $72.392^{\pm 6.687}$   & $24.810^{\pm 0.573}$ \\
        ``play the violin''             & $20/20$ & $111.449^{\pm 20.439}$ & $23.933^{\pm 1.169}$ \\
        \bottomrule
    \end{tabular}
    }
\end{table}
The framework achieved a 100\% success rate across all evaluated commands, corresponding to 80/80 successful trials in total, which demonstrates reliable real-world deployment. The local joint tracking error ($E_\text{mpjpe}$) remains within 22--33~mm, indicating precise joint-level control. The global tracking error ($E_\text{g-mpjpe}$) is higher for high-momentum motions such as ``punch'' due to root translation drift, but it remains within acceptable bounds and does not compromise semantic fidelity or physical stability.

\section{DISCUSSION AND CONCLUSION}
ECHO introduces an edge-cloud architecture for language-directed humanoid control, bypassing onboard constraints by decoupling diffusion generation from high-frequency RL tracking. Unitree G1 deployments validate that our 38D interface ensures dynamically feasible execution. To address current network and environmental limitations, future work will integrate visual feedback to evolve ECHO into an obstacle-aware Vision-Language-Action (VLA) architecture , ultimately providing a scalable, retargeting-free foundation for agile control.



\bibliographystyle{ieeetran.bst} 

\bibliography{main}

\section*{APPENDIX}
\label{sec:appendix}

This appendix details the evaluation metrics, reward formulations, domain randomization, and network hyperparameters.

\subsection{Evaluation Metrics}
\label{sec:metrics}

\noindent\textbf{Feature Extraction (MoCLIP).}
Since the standard HumanML3D evaluator~\cite{guo2022generating} is incompatible with our 38D format, we train a dedicated \textbf{MoCLIP} \cite{jia2025lumalowdimensionunifiedmotion} model: a 4-layer Transformer motion encoder (768-dim, 8 heads) aligned with a frozen CLIP ViT-L/14 text encoder~\cite{radford2021learning} via contrastive learning on the retargeted HumanML3D corpus. All experiments target the Unitree G1 EDU variant with 29 actuated DoF (hip $\times$6, knee $\times$2, ankle $\times$4, waist $\times$3, shoulder $\times$6, elbow $\times$4, wrist $\times$4; dexterous hands excluded).

\begin{itemize}
    \item \textbf{FID~\cite{heusel2017gans}:} Fréchet distance between Gaussians fitted on generated and ground-truth MoCLIP embeddings. Lower is better.
    \item \textbf{R-Precision:} Top-1/2/3 retrieval accuracy of a generated motion against 32 text candidates in MoCLIP space. Higher is better.
    \item \textbf{Diversity:} Average pairwise Euclidean distance over 300 randomly sampled motion pairs.
    \item \textbf{MM. Dist.~(Multimodal Distance):} Assesses the semantic alignment between generated motions and corresponding texts. Lower MM Dist. indicates better cross-modal correspondence.
\end{itemize}

\smallskip
\noindent\textbf{Robot-Centric Metrics (Proposed).}
We introduce two metrics assessing physical executability, which standard text-to-motion benchmarks omit.

\begin{itemize}[
\setlength{\itemsep}{0pt}
\setlength{\parsep}{0pt}
\setlength{\topsep}{0pt}
]
    \item \textbf{Motion Safety Score (MSS):}
    Measures compliance with Unitree G1 hardware constraints: joint position soft limits (90\% of range), velocity ($\pm 10$\,rad/s), and acceleration (100\,rad/s$^2$). For constraint $k$, a normalized violation $\overline{v}_k \geq 0$ is mapped to a sub-score $S_k = \exp(-100\cdot\overline{v}_k)$. The per-motion score is:
    \begin{equation}
        \text{MSS}_i = S_{\text{pos}}^{0.5} \cdot S_{\text{vel}}^{0.3} \cdot S_{\text{acc}}^{0.2}, \quad
        \text{MSS} = \tfrac{1}{N}\textstyle\sum_i \text{MSS}_i \in [0,1].
    \end{equation}

    \item \textbf{Root Trajectory Consistency (RTC):}
    Measures whether the generated root path matches the ground-truth in shape and travel extent. A Shape Score $S_{\text{shape}}$ compares arc-length-reparameterized waypoints ($K=50$, $\sigma=0.35$); an Extent Score $S_{\text{extent}}$ compares total arc lengths ($\sigma=0.8$):
    \begin{equation}
        \text{RTC}_i = S_{\text{shape}}^{0.7} \cdot S_{\text{extent}}^{0.3}, \quad
        \text{RTC} = \tfrac{1}{N}\textstyle\sum_i \text{RTC}_i \in [0,1].
    \end{equation}
\end{itemize}

\subsection{Reward Functions}
The detailed weight configurations are presented in Table \ref{tab:reward_weights}. All regularization terms are computed as penalties and scaled by their absolute weights.

\begin{table}[htbp]
\centering
\footnotesize
\renewcommand{\arraystretch}{0.85}
\caption{Reward Function Configurations}
\label{tab:reward_weights}
\begin{tabular}{lc}
\toprule
\textbf{Reward Term} & \textbf{Weight} \\
\midrule
\multicolumn{2}{c}{\textit{Motion Tracking}} \\
\midrule
Root Position Tracking & 0.5 \\
Root Rotation Tracking & 0.5 \\
Root Lin / Ang Velocity & 1.0 / 1.0 \\
Keypoint Tracking & 1.0 \\
Upper / Lower Keypoint Tracking & 0.5 / 0.5 \\
Joint Position Tracking & 1.0 \\
Joint Velocity Tracking & 0.5 \\
\midrule
\multicolumn{2}{c}{\textit{Locomotion Regularization}} \\
\midrule
Survival & 3.0 \\
Feet Air Time Ref & 5.0 \\
Feet Air Time Dense & 1.0 \\
Joint Velocity Penalty & $5.0 \times 10^{-4}$ \\
Joint Acceleration Penalty & $2.0 \times 10^{-8}$ \\
Action Rate Penalty & 0.01 \\
Joint Position Limits & 1.0 \\
Joint Torque Limits & 0.01 \\
\bottomrule
\end{tabular}
\end{table}

\subsection{Domain Randomization}
To facilitate zero-shot sim-to-real transfer, we apply domain randomization to physical parameters (sampled uniformly within bounds) as detailed in Table \ref{tab:domain_randomization}.

\begin{table}[htbp]
\centering
\footnotesize
\renewcommand{\arraystretch}{0.85}
\caption{Domain Randomization Parameters}
\label{tab:domain_randomization}
\begin{tabular}{lc}
\toprule
\textbf{Parameter} & \textbf{Randomization Range} \\
\midrule
Pelvis \& Torso Mass & $[0.9\times, 1.1\times]$ \\
Pelvis \& Torso COM Offset & $[-0.02, 0.02]$ m \\
Ankle Static Friction & $[0.5\times, 1.5\times]$ \\
Ankle Solref Time Constant & $[0.015, 0.03]$ s \\
Ankle Solref Damping Ratio & $[0.5, 2.0]$ \\
Joint Offset Error & $[-0.01, 0.01]$ rad \\
Motor Stiffness / Damping & $[0.8\times, 1.2\times]$ \\
Motor Armature & $[0.75\times, 1.25\times]$ \\
\bottomrule
\end{tabular}
\end{table}

\subsection{Network Architecture and Training Setup}
\textbf{Network Architecture:} The Teacher network encodes privileged information into a 256-dim latent vector using an MLP $[512, 256]$. The Student Adaptation Module (MLP: $[512, 512, 256]$) predicts this latent using proprioceptive history. Both Actor and Critic use MLPs $[1024, 512, 512]$ with LayerNorm and Mish activations. To encourage proper exploration, initial noise scales are explicitly set based on joint types: 1.5 for hips/knees/waist, 1.2 for shoulders/elbows, and 1.0 for wrists/ankles. Table \ref{tab:ppo_hyperparameters} summarizes the hyperparameters.

\begin{table}[htbp]
\centering
\footnotesize
\renewcommand{\arraystretch}{0.85}
\caption{PPO Hyperparameters \& Details}
\label{tab:ppo_hyperparameters}
\begin{tabular}{lc}
\toprule
\textbf{Parameter} & \textbf{Value} \\
\midrule
Total Training Frames & $8 \times 10^9$ \\
PPO Epochs / Minibatches & 5 / 8 \\
Learning Rate & $5 \times 10^{-4}$ \\
Target KL Divergence & 0.01 \\
Discount Factor ($\gamma$) / GAE ($\lambda$) & 0.99 / 0.95 \\
Clip Range & 0.2 \\
Entropy Coefficient & Decays 0.01 $\rightarrow$ 0.0025 \\
Distillation Weight ($\lambda_{\text{reg}}$) & 0.2 \\
Action Delay Buffer & 3 steps \\
Action Scaling (Upper / Lower) & 1.0 / 0.5 \\
Symmetry Loss & Enabled \\
\bottomrule
\end{tabular}
\end{table}

\end{document}